\documentclass[10pt,twocolumn,letterpaper]{article}

\usepackage{cvpr}              

\usepackage[dvipsnames]{xcolor}

\usepackage{graphicx}
\usepackage{amsmath}
\usepackage{amssymb}
\usepackage{booktabs}
\usepackage{makecell}
\usepackage{multirow}
\usepackage{cuted}
\usepackage{caption, subcaption, overpic, textpos}
\usepackage{enumitem}
\usepackage{color, colortbl}
\usepackage{comment}
\definecolor{LightCyan}{rgb}{0.88,1,1}

\newcommand{\newtitle}{Improving Distant 3D Object Detection Using 2D Box Supervision\xspace}

\definecolor{cvprblue}{rgb}{0.21,0.49,0.74}
\usepackage[pagebackref,breaklinks,colorlinks,citecolor=cvprblue]{hyperref}
\hypersetup{citecolor=[RGB]{119,185,0}}

\def\paperID{3152} 
\def\confName{CVPR}
\def\confYear{2024}

\newcommand{\myparagraph}[1]{\vspace{3pt}\noindent{\bf #1}}
\newcommand{\tablestyle}[2]{\setlength{\tabcolsep}{#1}\renewcommand{\arraystretch}{#2}\centering\footnotesize}

\title{\newtitle}

\author{
Zetong Yang$^{1}$\thanks{Work done during internship / affiliation with NVIDIA.}~~
Zhiding Yu$^{2}$\thanks{Corresponding author: \href{mailto:zhidingy@nvidia.com}{zhidingy@nvidia.com}}~~
Chris Choy$^{2}$~
Renhao Wang$^{3\,*}$~
Anima Anandkumar$^{4\,*}$~
Jose M. Alvarez$^{2}$
\\
\vspace{-0.8mm}
\\
$^{1}$CUHK~~~~~~$^{2}$NVIDIA~~~~~~$^{3}$UC Berkeley~~~~~~$^{4}$Caltech
}

\begin{document}
\maketitle

\begin{strip}
  \vspace{-46pt}
  \centering
  \includegraphics[width=1.0\linewidth]{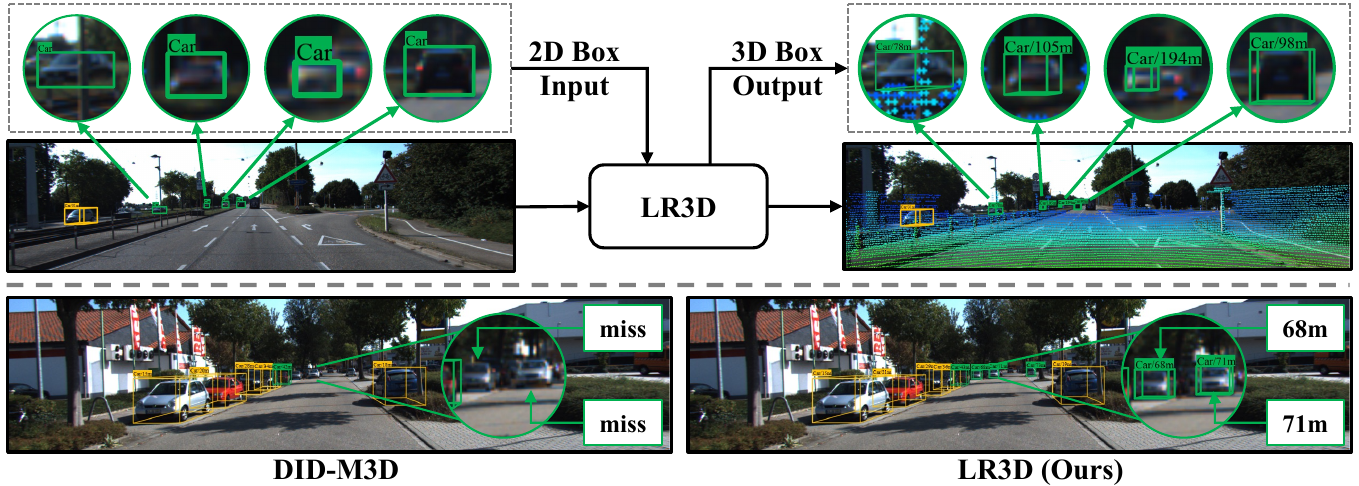}\\
  \vspace{-9pt}
  \captionof{figure}{\textbf{Upper:} Without 3D annotations beyond 40m, LR3D enables the predictions of 3D boxes for extremely distant objects over 200m (right, in \textcolor[rgb]{0,0.69,0.31}{\textbf{green}}) based on the inputs of image and 2D box (left). \textbf{Lower:} Existing methods fail to detect 3D objects beyond the 3D supervision range, \eg, 40m for this case. With LR3D, these remote missing objects are well detected.
  }
  \vspace{-5pt}
  \label{fig:teaser}
\end{strip}

\begin{abstract}
\vspace{-.1in}
Improving the detection of distant 3d objects is an important yet challenging task. For camera-based 3D perception, the annotation of 3d bounding relies heavily on LiDAR for accurate depth information. As such, the distance of annotation is often limited due to the sparsity of LiDAR points on distant objects, which hampers the capability of existing detectors for long-range scenarios. We address this challenge by considering only 2D box supervision for distant objects since they are easy to annotate. We propose LR3D, a framework that learns to recover the missing depth of distant objects. LR3D adopts an implicit projection head to learn the generation of mapping between 2D boxes and depth using the 3D supervision on close objects. This mapping allows the depth estimation of distant objects conditioned on their 2D boxes, making long-range 3D detection with 2D supervision feasible. Experiments show that without distant 3D annotations, LR3D allows camera-based methods to detect distant objects (over 200m) with comparable accuracy to full 3D supervision. Our framework is general, and could widely benefit 3D detection methods to a large extent.
\vspace{-2mm}
\end{abstract}

\vspace{-.1in}
\section{Introduction}

Camera-based 3D object detection \cite{wang2021fcos3d,li2022bevformer,li2023bevdepth,liu2022petr,yang2023bevformerV2,peng2022did,wang2022mvfcos3d++} is an important task in autonomous driving aiming to localize and classify objects in 3D space with monocular or multi-view image input. Detecting distant objects with camera input is both important and challenging. On one hand, the ability to detect objects at distance is needed for planning, especially for highway scenarios. According to \cite{brakingdistWiki}, at 60 miles/hour, the typical stopping distance for a vehicle is 73 meters and it grows significantly under harsh road conditions such as having wet surface \cite{R4D2022Li,longstoppingdistances,blanco2005visual}. On the other hand, the detection range of camera-based methods depends heavily on the distance range of 3D annotations. For example, these methods work well within the distance range with abundant 3D annotations (\eg, $\sim$70 meters on KITTI \cite{KITTIDATASET1} and nuScenes \cite{nuscenes2019}), but often fail beyond this range where 3D annotations are missing. This indicates the importance to improve distant 3D object detection with longer 3D annotation range.

Though of great importance, increasing the range of 3D annotation is not easy. One main challenge is the sparsity of LiDAR points at longer distances. LiDAR is typically the main reference for axial depth information in 3D annotation. For distant objects with few or even no LiDAR points, labeling their 3D bounding boxes is an ill-posed, noisy, and time-consuming task for human annotators. For this reason, distant 3D box annotations are rare, and most datasets \cite{KITTIDATASET1,nuscenes2019,Waymo} provide 3D annotations around 70 meters. A few datasets \cite{Argoverse2,cityscape3d} provide far-away annotations. They obtain long-range 3D annotations either by using costly LiDAR sensors with very long sensing ranges or by taking significant efforts in annotating 3D boxes with consistency among sequences of stereo images. Both of these strategies are more expensive and less scalable, compared to drawing precise 2D boxes on images. Thus, in this paper, we explore an efficient camera-based approach to achieve high-quality long-range 3D object detection, which uses only 2D annotations for long-range objects of which the 3D annotations are hardly available due to sparse or no interior LiDAR points.

\textbf{Our approach:}
We present LR3D, a camera-based detection framework that detects distant 3D objects using only 2D bounding box annotations. The core of LR3D is an Implicit Projection Head (IP-Head) design that can be plugged into existing camera-based detectors and enables them to effectively predict 3D bounding boxes at all ranges, using the supervision from close 2D/3D and distant 2D box labels.

IP-Head learns to generate specific mapping for each instance from their 2D bounding box prediction to the corresponding depth, and thus is capable of estimating depth for distant objects relying on 2D supervision only. We also design a projection augmentation strategy to force the generated implicit function to correctly model the mapping between 2D boxes and 3D depth of target instances and estimate different depth outputs if the 2D box input changes.

We also notice issues in evaluating camera-based detectors for long-range 3D detection. Existing evaluation metrics \cite{nuscenes2019,KITTIDATASET1}, based on mean average precision (mAP) with fixed thresholds, neglect the inaccuracy of depth estimation of distant objects from cameras and lead to meaningless numbers even for state-of-the-art methods, \eg, only 0.43\% mAP of DID-M3D \cite{peng2022did} on KITTI Dataset for objects farther than 40m. This motivates us to design a novel metric, Long-range Detection Score (LDS), which sets a dynamic threshold for judging TP prediction regarding associated ground truth depth, for informative quatitative comparison.

We conduct experiments on five popular 3D detection datasets including KITTI \cite{KITTIDATASET1}, nuScenes \cite{nuscenes2019}, and Waymo \cite{Waymo} for objects within 80m, as well as Cityscapes3D \cite{cityscape3d} and Argoverse 2 \cite{Argoverse2} for objects farther than 200m. We remove the 3D annotations of farther objects in the train set and evaluate the performance on 3D objects at all ranges.

Experiments show that, with LR3D, state-of-the-art detectors gain significant improvement in detecting distant objects without 3D annotation, \ie, 14.8\% improvement of DID-M3D \cite{peng2022did} on KITTI \cite{KITTIDATASET1}, 15.9\% improvement of BEVFormer \cite{li2022bevformer} on nuScenes \cite{nuscenes2019}, 34.5\% and 14.17\% improvement of FCOS3D \cite{wang2021fcos3d} on Cityscapes3D and Argoverse 2, 7.09\% improvement of MV-FCOS3D++ \cite{wang2022mvfcos3d++} on Waymo. With LR3D, these methods yield competitive performance even compared to fully 3D supervised counterparts. Notably, LR3D enables them to detect extremely distant 3D objects as shown in Figure \ref{fig:teaser}.

\section{Related Work}
\vspace{-.1in}
\myparagraph{LiDAR-based Detectors.}
LiDAR methods detect 3D objects from point clouds. According to inputs, these methods can be divided into point, voxel, and range-view detectors. Point methods \cite{shi2018pointrcnn,FPOINTNET,yang3DSSD20,yang2019std,yang2018ipod} make 3D predictions from raw point clouds \cite{POINTNET,POINTNET2,PointCNN,PointTransformer,eq2022yang,yang2020cn}.
Voxel methods \cite{yin2021center,3DMAN,lang2018pointpillars,PVRCNN} transform point clouds into voxels and extract features by convolutions \cite{sparseconv,sparseconvold,Minkowski,yan2018second,YangLU18,jiang2023msp}.
Other methods project point clouds into range-view \cite{pei2020rangeconv,pei2021rsn,bo2016rangedet,fan2021rangedet,cai2021ttp} and process them as images \cite{ResNet}. 
Albeit impressively performed, LiDAR methods are 3D-label-greedy, and limited by the perspective range of LiDAR sensors, \ie, usually fail for areas with few or no LiDAR points.

\vspace{-0.03in}
\myparagraph{Camera-based Detectors.}
Camera-based methods do 3D detection from images. Monocular methods predict 3D boxes in single image directly \cite{wang2021fcos3d,wang2021pgd,brazil2019m3drpn,andrea2019mono,3dgck}. Stereo methods based on multi-view images formulate a 3D volume \cite{chen2020dsgn,Guo_2021_ICCV,chen2019mvs}. Recent methods \cite{li2022bevformer,liu2022bevfusion,rukhovich2022imvoxelnet,CaDDN,li2022uvtr,philion2020lift,yang2023vidar,liu2023fully} build a bird-eye-view (BEV) representation for detection. 

The advantages of Camera-based methods lie in the unbounded perception range of cameras, which makes them suitable for distant 3D detection. Some methods, \eg, Far3D \cite{jiang2023far3d}, are designed for long-range 3D detection, but still heavily rely on abundant high-quality distant 3D annotations, which are hard to collect.
Some datasets \cite{Argoverse2,cityscape3d} provide distant annotations. However, their labeling is time-consuming and requires huge human efforts to take hints from multi-modal sensors and temporal consistency to label distant objects with few or no interior LiDAR points.
The workload of the labeling process and the 3D-label-greedy character of existing detectors limit the applications of long-range 3D detection.
Instead, we propose to detect distant 3D objects without long-range 3D box supervision. It is the simplest setting and is practical for scalability.

\vspace{-0.03in}
\myparagraph{Object Distance Estimation.}
Another related topic is monocular distance estimation, which estimates the object distance from an RGB image. SVR \cite{gokcce2015vision} and DisNet \cite{haseeb2018disnet} estimate object distance through pixel height and width. They base on an assumption of projected 2D box size determined by the distance of object only. However, other factors, like size and orientation, also affect the projected 2D box size. 

Method of \cite{zhu2019learning} develops a FastRCNN \cite{FASTRCNN} pipeline for end-to-end distance estimation by directly regressing the distance from RoI features. R4D \cite{R4D2022Li} utilizes reference objects for further improvements. These methods are limited by their requirements of abundant annotations. Given the difficulty in labeling distant 3D objects, in this paper, we develop a simpler 2D setting that even achieves competitive performance to their 3D supervised counterparts.

\begin{figure}[t]
  \vspace{-0.2in}
  \centering
  \includegraphics[width=1.0\linewidth]{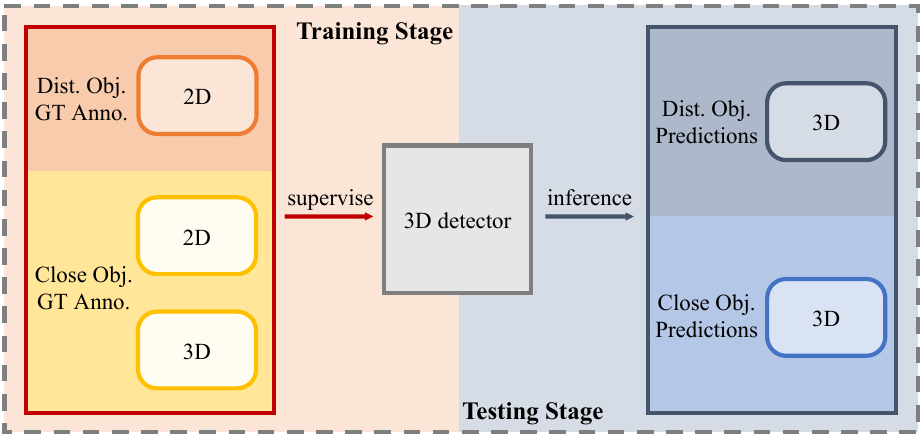}\\
  \vspace{-0.1in}
  \caption{Illustration of LR3D which detects 3D boxes for both close and distant objects using the supervision of close 2D/3D and distant 2D bounding box annotations.}
  \label{fig:ip_lr3d}
  \vspace{-0.1in}
\end{figure}
\begin{table}[t]
	\centering \addtolength{\tabcolsep}{1pt}
	\footnotesize
	\begin{tabular}{ c | c | c | c }
		\hline
	    \multicolumn{1}{c|}{ \multirow{2}{*}{\makecell[c]{Distant 3D\\Groundtruth?}}} &
	    \multicolumn{1}{c|}{ \multirow{2}{*}{Location Err.}} & 
	    \multicolumn{1}{c|}{ \multirow{2}{*}{Size Err.}} & 
	    \multicolumn{1}{c}{ \multirow{2}{*}{Orientation Err.}} \\
	    & & &\\
		\hline
		$\surd$ & 0.09 & 0.21 & 0.37 \\
		- & \bf 0.34 (+0.25) & 0.23 (+0.02) & 0.41 (+0.04) \\
		\hline
	\end{tabular}
	\vspace{-0.1in}
	\caption{Performance comparison between FastRCNN3D trained with and without distant 3D supervision.}
	\label{tab:main_obstacle}
	\vspace{-0.2in}
\end{table}
\section{Method}
LR3D is a long-range 3D detection framework that detects 3D bounding boxes, including locations, sizes, and orientations, of distant objects using only their 2D supervision (Figure \ref{fig:ip_lr3d}). In this section, we first analyze the main challenge of directly utilizing existing 3D detectors on LR3D in Section \ref{sec:lr3d} and then introduce our solution in Section \ref{sec:iphead}.

\subsection{Analysis} \label{sec:lr3d}
Albeit important and critical, long-range 3D object detection is seldom explored. One difficulty is to obtain sufficient high-quality 3D bounding box annotations for distant objects. We discard this high-human-labor-cost setting and propose a new framework, LR3D, to predict 3D boxes for distant objects only from their 2D image supervision.

\myparagraph{Can Existing Methods Use only 2D Supervision?}
We first analyze if existing methods can use only 2D supervision for detecting distant 3D objects. To this end, we use FastRCNN3D, a FastRCNN-like \cite{FASTRCNN} 3D detector where the original head for 2D detection is replaced by the FCOS3D head \cite{wang2021fcos3d} for 3D detection. For training, we first manually assign objects over 40m as distant objects and remove their 3D annotations. Then, we train FastRCNN3D with close 2D/3D and distant 2D box labels as Figure~\ref{fig:ip_lr3d}. At test time, we use the 2D ground truth of distant objects as proposals, predict their 3D boxes, and evaluate their errors in location, size, and orientation. Note that we use relative distance as location error, IoU with aligned location and orientation as size error, and absolute difference as orientation error.

\begin{figure}[t]
  \vspace{-0.2in}
  \centering
  \includegraphics[width=1.0\linewidth]{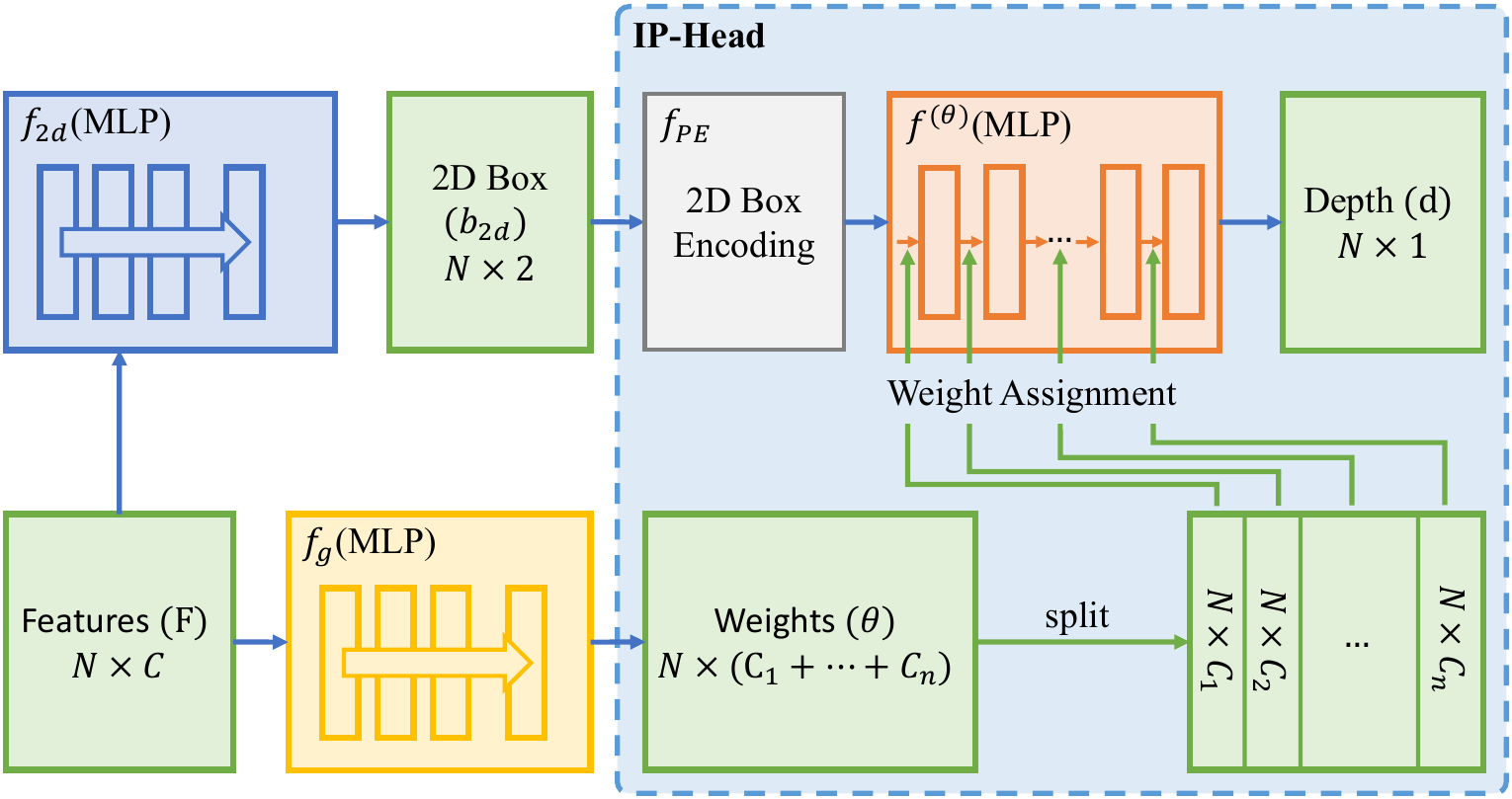}\\
  \vspace{-0.1in}
  \caption{Illustration of IP-Head. We use an MLP $f^{(\theta)}$ to fit the implicit function from 2D box to 3D depth, of which the weights $\theta$ are dynamically determined by instance features including information of size and orientation.
  }
  \label{fig:ip_impl}
  \vspace{-0.1in}
\end{figure}

Table~\ref{tab:main_obstacle} compares errors between the LR3D configuration and traditional full supervision. As shown, errors in size and orientation are comparable. The main gap comes from the location error, in which depth estimation plays the essential role. Due to the lack of depth information in 2D box supervision, directly regressing the depth as previous methods cannot produce accurate depth prediction for distant objects and leads to a significant drop. Therefore, we seek a new method to implicitly estimate the depth.

\begin{figure*}[t]
    \vspace{-0.2in}
	\centering
	\includegraphics[width=1.0\linewidth]{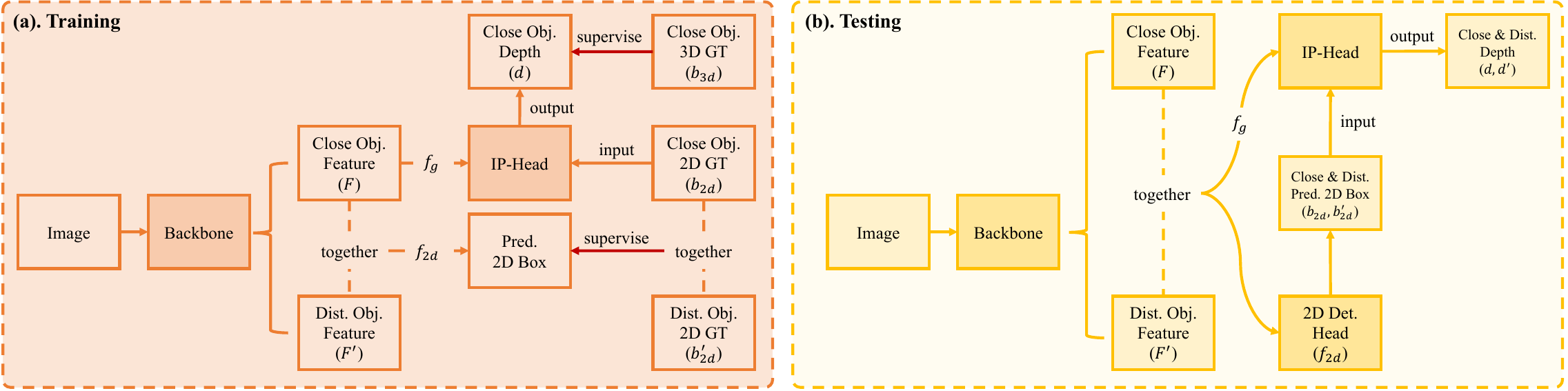}\\
	\vspace{-0.1in}
	\caption{Illustration of the training and testing pipeline of IP-Head. \textbf{(a). Training:} During training, we use 2D/3D annotation pairs of close objects to supervise $f_g$ to generate dynamic weights of MLP $f^{(\theta)}$ which models the transformation of target 3D object from 2D box to corresponding depth in Eq. \eqref{eq:ipl_f_inverse_sim}. \textbf{(b). Testing:} During testing, we use a 2D detection head (2D Det. Head) $f_{2d}$ to generate 2D detection results for all objects. They are then transferred to corresponding depth by IP-Head.
	}
	\label{fig:iphead_main}
	\vspace{-0.2in}
\end{figure*}

\subsection{Implicit Projection Head} \label{sec:iphead}
Here, we introduce Implicit Projection head (IP-Head), our proposal to estimate depth of distant objects using only 2D supervision. Given a 3D object with fixed depth, size, and orientation \footnote{The mentioned orientation is the relative one to the camera, i.e., the observed orientation on the image. We refer the readers to FCOS3D \cite{wang2021fcos3d} and Stereo-RCNN \cite{stereorcnn} for more details on obtaining relative orientation.}, through the camera calibration matrix, it is easy to obtain the corresponding projected 2D bounding box (described by its width $w_{2d}$ and height $h_{2d}$) on the target image. We use a function $f$, determined by the calibration matrix, to indicate the mapping between depth ($d$), size ($s$), orientation ($o$) and 2D box ($b_{2d} = (w_{2d}, h_{2d})$) as
\begin{equation} \label{eq:ipl_f}
	\begin{aligned}
		f(d, s, o) = b_{2d}.
	\end{aligned}
\end{equation}
Eq. \eqref{eq:ipl_f} shows the ubiquitous relation between $d$ and $b_{2d}$ if the object size $s$ and orientation $o$ are fixed -- for objects with the same size and orientation, the further these objects locate, the smaller their projected 2D boxes on image are. Inspired by this fact, we ask if it is possible to estimate the inverse function $f^{-1}$ to transfer the 2D bounding box to the corresponding depth conditioned by $s$ and $o$, formulated as
\begin{equation} \label{eq:ipl_f_inverse}
	\begin{aligned}
		f^{-1}(b_{2d} | s, o) = d.
	\end{aligned}
\end{equation}
With the power of neural networks to fit complicated functions, we utilize a small-size network with a multi-layer perceptron (MLP) to estimate the implicit inverse function $f^{-1}$. For simplicity, we use $f^{(\theta)}$ to represent this network, of which the parameter weights are represented as $\theta$.

Since the implicit inverse function $f^{-1}$ depends on the size and orientation of the specific 3D objects, $f^{(\theta)}$ should also be different across multiple objects, which means weights $\theta$ should be dynamic.

\begin{figure}[t]
  \vspace{-0.05in}
  \centering
  \includegraphics[width=0.95\linewidth]{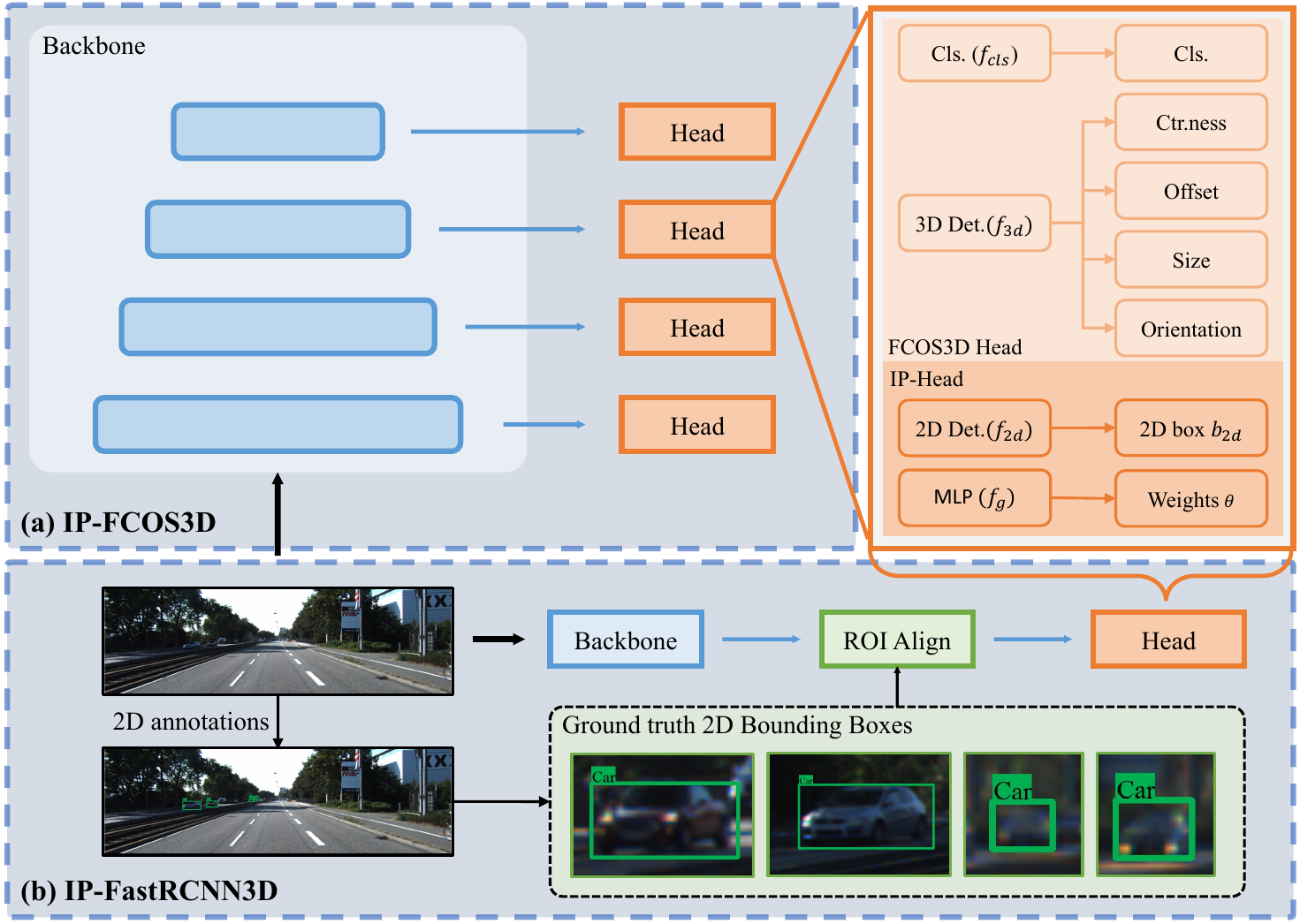}\\
  \vspace{-0.1in}
  \caption{Illustration of deploying IP-Head to monocular 3D detectors: \textbf{(a)} FCOS3D; \textbf{(b)} FastRCNN3D.}
  \label{fig:ip_networks}
  \vspace{-0.2in}
\end{figure}

With these considerations, rather than utilizing a shared $\theta$ for all objects, we use a trainable MLP $f_g$ to generate a set of dynamic weights $\theta_i$ according to the features $F_i$ of each object $i$. We then utilize those $\theta_i$ as the weights of network $f^{(\theta)}$ to estimate the corresponding depth of the $i$-th 2D box. This process generates the specific \textbf{I}mplicit inverse function of each object to \textbf{P}roject its 2D box to 3D depth. We illustrate this procedure in Figure \ref{fig:ip_impl}, in which ``2D Box Encoding'', $f_{\text{PE}}$, is a positional encoding function \cite{AttentionIsAllYouNeed} to encode 2-channel 2D box descriptors (with width $w_{2d}$ and height ${h_{2d}}$) into informative high-dimensional features and $f_{2d}$ is a 2D detection network for predicting 2D bounding boxes on the image. The overall procedure is formulated as
\begin{equation} \label{eq:ipl_f_inverse_sim}
	\begin{aligned}
	    d_i = f^{(f_g(F_i))}(f_{\text{PE}}(b_{{2d}_i})),
	\end{aligned}
\end{equation}
where $f_g$ estimates the weights of $f^{(\theta_i)}$ from instance features $F_i$ to transfer the $i$-th 2D bounding boxes to its corresponding depth.
Its condition information, including size $s_i$ and orientation $o_i$, is included in feature $F_i$, from which information can be obtained like that of \cite{wang2021fcos3d,MonoFlex,lu2021geometry}.

\begin{figure}[!t]
  \vspace{-0.05in}
  \centering
  \includegraphics[width=1.0\linewidth]{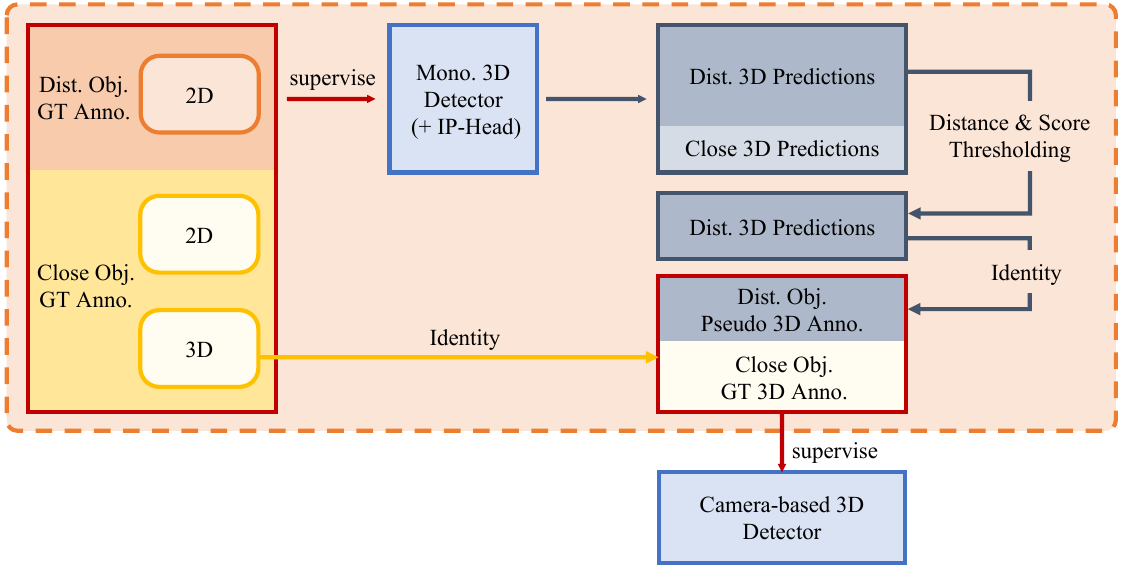}\\
  \vspace{-0.1in}
  \caption{Illustration of extending IP-Head to all camera-based 3D detectors through a teacher-student pipeline.
  }
  \label{fig:ip_faraway_teacher}
  \vspace{-0.2in}
\end{figure}
During training, we make use of 2D/3D annotation pairs of close objects to supervise IP-Head for obtaining a reliable dynamic weight generator $f_g$. Specifically, for a close object, after obtaining its corresponding dynamic weights, we transfer its 2D ground truth box for depth prediction by IP-Head, and optimize $f_g$ by computing the loss between the predicted depth and the 3D annotation. 

During inference, we first use the backbone network to extract instance features $F^\prime_i$. Then, we extract dynamic weights using $f_g$ and use a detection network $f_{2d}$, supervised by 2D bounding box ground truth, to predict the associated 2D bounding box $b^\prime_{{2d}_i}$ as 	    $b_{{2d}_i}^\prime = f_{2d}(F^\prime_i)$. Finally, we use Eq. \eqref{eq:ipl_f_inverse_sim} to obtain depth predictions for each instance in the image. The overall process is illustrated in Figure \ref{fig:iphead_main}.

Using neural networks to directly estimate a hard-to-optimize function was successful in other research areas. For example, NeRF \cite{mildenhall2020nerf} uses MLP networks to learn a shared mapping from positions to colors and densities; classification models use a shared CNN model to learn the function from pixels to classification confidence. In contrast to previous methods, we first study the estimation of transformation function from 2D boxes on images to 3D depths. Moreover, instead of learning a shared mapping, IP-Head generates dynamic mappings from 2D box sizes to 3D depth according to the 3D size and orientation of target objects, which equips IP-Head with great generalization ability and potential applied to complex autonomous driving scenarios.

\myparagraph{Projection Augmentation.}
The implicit inverse function $f^{(\theta)}$ needs to model the relation between the 2D box and corresponding depth, and generate different depth predictions based on 2D box input. To further improve effectiveness, we also propose an augmentation strategy, called projection augmentation. The idea is to generate more depth $d$ and 2D box $b_{2d}$ pairs for each close object during training, so as to enable $f_g$ to estimate a more accurate $b_{2d}$-$d$ relation.

These extra $b_{2d}$-$d$ training pairs come from Eq. \eqref{eq:ipl_f}. Given an object with fixed size and orientation, we randomly choose different depth values $d^\ast$, calculate their corresponding 2D boxes $b_{2d}^\ast$ through Eq. \eqref{eq:ipl_f}, and utilize these augmented $b_{2d}^\ast$-$d^\ast$ pairs, along with the ground truth $b_{2d}$-$d$ pair, to train the IP-Head for higher performance.

\myparagraph{Long-range Teacher}
The proposed IP-Head can be used in existing monocular 3D detectors, like FCOS3D \cite{wang2021fcos3d}, to boost their performance in LR3D. As illustrated in Figure \ref{fig:ip_networks}, only two additional branches, i.e., a 2D detection branch $f_{2d}$ and a weight generation MLP $f_g$, are needed to utilize IP-Head in FCOS3D. Apart from monocular methods, BEV methods become popular due to their strong performance and multi-tasking ability. We extend IP-Head to BEV methods to alleviate their demand of 3D annotations.

Our solution is to utilize a monocular method equipped with IP-Head as the detector of the LR3D model which then serves as the long-range teacher to generate pseudo distant 3D annotations. With close 2D/3D annotations and distant 2D annotations, we first apply LR3D to generate 3D predictions for distant objects. Then, we treat these predictions as pseudo 3D box labels, which are, together with close 3D ground truth, taken as the whole supervision to train BEV methods. Our pipeline is shown in Figure \ref{fig:ip_faraway_teacher}. Experiments show that, it works decently for various 3D detectors on long-range objects without distant 3D box annotation.

\vspace{-.05in}
\section{Long-range Detection Score}
\vspace{-.05in}
Existing detection metrics are built upon mAP based on either IoU \cite{KITTIDATASET1,Waymo} or absolute distance error \cite{nuscenes2019}. They use fixed IoU or error thresholds as criterion to calculate AP. 

However, the fixed threshold is not suitable for distant objects due to its neglect of the increasing error in depth estimation when objects go farther. For example, though DID-M3D \cite{peng2022did} is the state-of-the-art camera-based detector, it only achieves 0.43\% mAP under fixed IoU criterion on objects over 40m. Fixed threshold metrics hardly bring informative results. In fact, further objects should have a larger tolerance to closer ones, since estimating their depth is more ill-posed. Similar ideas were taken in tasks like dense depth estimation \cite{Ranftl2021} and object distance estimation \cite{R4D2022Li}. They all prefer relative distance errors as their metrics. 

To address this limitation, we introduce Long-range Detection Score (LDS), a new metric for long-range 3D object detection. LDS is based on the widely applied nuScenes Detection Score (NDS) \cite{nuscenes2019} and it is defined as 
\begin{equation} \label{eq:lds}
	\begin{aligned}
		\text{LDS}\! =\! \frac{1}{6}[3\text{mAP}\! +\! \text{Rec}\!\times\!\sum_{\text{mTP} \in \text{m}\mathbb{TP}}(1 - \text{min}(1, \text{mTP}))],
	\end{aligned}
\end{equation}
\noindent where Rec is the recall rate and mTP represents the mean True Positive metric.

LDS improves NDS in two main aspects: the criterion of mAP and the multiplication of Rec and summation of mTP. 
\myparagraph{Improvements on mAP.}
In LDS, we compute mAP based on the relative distance error of
\begin{equation} \label{eq:rde}
	\begin{aligned}
		\text{Rel. Dist. Err.} = \frac{\| P_{c} - G_{c}  \|}{G_{d}},
	\end{aligned}
\end{equation}
where $P_c$, $G_c$ and $G_d$ represent the center of predicted 3D box, center of ground truth 3D box and the distance of ground truth 3D box towards ego vehicle, respectively.
Predictions with a relative error smaller than a threshold $r$ are counted as true positive, and false positive otherwise, for computing AP. We choose 4 thresholds $\mathbb{R}=\{0.025, 0.05, 0.1, 0.2\}$ and take average over these thresholds and the class set $\mathbb{C}$. Finally, we obtain mAP as
\begin{equation} \label{eq:mAP}
	\begin{aligned}
		\text{mAP} = \frac{1}{|\mathbb{C}| |\mathbb{R}|}\sum_{c \in \mathbb{C}} \sum_{r \in \mathbb{R}} \text{AP}_{c,r}.
	\end{aligned}
\end{equation}

\myparagraph{Improvement on mTP.}
Also, we multiply the recall rate and mTP before adding to mAP. The mTP is utilized to measure errors on the location (mATE), size (mASE) and orientation (mAOE) for TP prediction, whose relative distance to the ground truth is smaller than $r=0.1$ during matching. mATE is computed as the relative distance, normalized by 0.1 to ensure range falling within 0 and 1. 

mASE and mAOE are the same as those in nuScenes Dataset \cite{nuscenes2019}. The intuition of multiplying the recall rate to the mTP is simple. The larger the recall rate is, the more predictions are involved in the statistics of mTP. 
Compared to simply setting a recall threshold \cite{nuscenes2019}, the multiplication improvement adjusts the weight of mTP to LDS according to its comprehensiveness, and thus brings a more informative quantitative result.
\section{Experiments}
We evaluate our method on five popular 3D detection datasets of KITTI \cite{KITTIDATASET1}, nuScenes \cite{nuscenes2019}, Cityscapes3D \cite{cityscape3d}, Waymo Open Dataset \cite{Waymo} and Argoverse 2 \cite{Argoverse2}, which have plenty of high-quality 3D annotations for both close and distant objects. \textbf{Due to the space limitation, we show the experimental results on Cityscapes3D, Waymo and Argoverse 2 datasets in the supplementary material.}

\begin{table*}[t]
    \vspace{-0.2in}
   \centering 
   \setlength{\tabcolsep}{3.8mm}{
   \footnotesize
   \begin{tabular}{l|c||cc||cc||cc}
       \hline
       \multicolumn{1}{c|}{ \multirow{2}{*}{Method}} &
       \multicolumn{1}{c||}{ \multirow{2}{*}{\makecell[c]{Distant 3D\\Groundtruth?}}} &
       \multicolumn{2}{c||}{Overall (0m-Inf)} & 
       \multicolumn{2}{c||}{Close (0m-40m)} & 
       \multicolumn{2}{c}{Distant (40m-Inf)} \\ \cline{3-8}
       \multicolumn{1}{c|}{} & 
       \multicolumn{1}{c||}{} & 
       LDS (\%) & mAP (\%) &
       LDS (\%) & mAP (\%) &
       LDS (\%) & mAP (\%) \\
       \hline
       \hline
       FCOS3D \cite{wang2021fcos3d} & $\surd$ & 48.0 & 44.7 & 49.9 & 47.5 & 38.2 & 32.6 \\
       \hline
       FCOS3D \cite{wang2021fcos3d} & \multicolumn{1}{c||}{ \multirow{2}{*}{-}} & 42.1 & 38.9 & 50.5 & 47.1 & 4.9 & 3.3 \\
       \makecell[l]{LR3D (IP-FCOS3D)} & {} & \bf 50.0 & \bf 46.7 & 52.1 & 49.4 & \bf 36.2 & \bf 31.0 \\
       \hline
       \hline
       \multicolumn{1}{l}{\multirow{1}{*}{\bf Long-range Teacher}} & \multicolumn{1}{l}{\multirow{1}{*}{}} & \multicolumn{1}{l}{\multirow{1}{*}{}} &
       \multicolumn{1}{l}{\multirow{1}{*}{}} & \multicolumn{1}{l}{\multirow{1}{*}{}} &
       \multicolumn{1}{l}{\multirow{1}{*}{}} & \multicolumn{1}{l}{\multirow{1}{*}{}} & \\
       \hline
       \hline
       ImVoxelNet \cite{rukhovich2022imvoxelnet} & $\surd$ & 44.8 & 45.2 & 48.1 & 48.4 & 26.2 & 24.7 \\
       \hline
       ImVoxelNet \cite{rukhovich2022imvoxelnet} & \multicolumn{1}{c||}{ \multirow{2}{*}{-}} & 40.3 & 39.9 & 48.6 & 48.1 & 4.4 & 4.1 \\
       +LR3D teacher & {} & \bf 44.9 & \bf 45.1 & 47.9 & 47.9 & \bf 26.9 & \bf 25.3 \\
       \hline
       \hline
       CaDDN \cite{CaDDN} & $\surd$ & 51.1 & 50.1 & 57.8 & 56.3 & 19.3 & 18.5 \\
       \hline
       CaDDN \cite{CaDDN} & \multicolumn{1}{c||}{ \multirow{2}{*}{-}} & 48.9 & 48.6 & 58.5 & 57.9 & 5.5 & 5.7 \\
       +LR3D teacher & {} & \bf 51.1 & \bf 50.6 & 57.6 & 56.8 & \bf 20.9 & \bf 19.8 \\
       \hline
       \hline
       MonoFlex \cite{MonoFlex} & $\surd$ & 55.5 & 52.4 & 57.7 & 55.2 & 40.6 & 34.8 \\
       \hline
       MonoFlex \cite{MonoFlex} & \multicolumn{1}{c||}{ \multirow{2}{*}{-}} & 50.1 & 47.3 & 59.3 & 56.1 & 8.4 & 6.6 \\
       +LR3D teacher & {} & \bf 54.3 & \bf 51.6 & 57.3 & 54.8 & \bf 35.8 & \bf 32.9 \\
       \hline
       \hline
       GUPNet \cite{lu2021geometry} & $\surd$ & 49.3 & 47.5 & 54.0 & 52.1 & 26.7 & 25.1 \\
       \hline
       GUPNet \cite{lu2021geometry} & \multicolumn{1}{c||}{ \multirow{2}{*}{-}} & 46.8 & 45.2 & 54.0 & 52.3 & 17.6 & 16.5 \\
       +LR3D teacher & {} & \bf 49.4 & \bf 47.8 & 54.2 & 52.8 & \bf 26.4 & \bf 24.4 \\
       \hline
       \hline
       DID-M3D \cite{peng2022did} & $\surd$ & 56.1 & 55.1 & 58.5 & 57.2 & 38.9 & 36.7 \\
       \hline
       DID-M3D \cite{peng2022did} & \multicolumn{1}{c||}{ \multirow{2}{*}{-}} & 52.5 & 51.4 & 58.9 & 57.8 & 24.2 & 22.5 \\
       +LR3D teacher & {} & \bf 56.3 & \bf 55.0 & 58.7 & 57.2 & \bf 39.0 & \bf 36.2 \\
       \hline
   \end{tabular}}
   \vspace{-0.1in}
   \caption{Comparison on state-of-the-art methods with and without IP-Head or LR3D teacher supervised by distant 2D ground truth only on the KITTI val dataset. Their fully supervised counterparts (with distant 3D ground truth) are also illustrated.
   }\label{tab:kitti_main}
   \vspace{-0.2in}
\end{table*}
\subsection{Results on KITTI}
\myparagraph{Data Preparation.}
KITTI Dataset \cite{KITTIDATASET1} provides high-quality 2D/3D annotations for objects in ``Car'', ``Pedestrian'' and ``Cyclist'' within the range of $\sim$80m. For extremely distant objects, due to lack of points, it only labels their 2D ground truth boxes on images and marks them as ``DontCare''. Given these labels, we accordingly conduct quantitative and qualitative evaluations on KITTI Dataset.

Quantitative evaluation is conducted on objects with 3D annotations. Specifically, based on the official provided 3D annotations, we mark annotations over 40m as distant ones following \cite{R4D2022Li}, only using their 2D labels for training. For those closer than 40m, we keep both 2D and 3D labels for training. We report LDS on both close and distant objects for statistical results. The split of train and val sets follows \cite{VOXELNET}. Due to the lack of ``Pedestrian'' and ``Cyclist'' labels beyond 40m, we only report results on class ``Car''.

Qualitative evaluation is designed for ``DontCare" objects which are extremely far away and without 3D annotations due to no LiDAR points. We visualize the 3D detection results of our model conditioned by their 2D ground truth boxes. Through this evaluation, we manifest the capacity of our model for extremely distant 3D detection.

\myparagraph{Model Setting.}
For quantitative evaluation, we utilize FCOS3D \cite{wang2021fcos3d} as our baseline for its simplicity and implementation platform. FCOS3D is implemented on the MMDetection3D \cite{mmdet3d2020} platform including multiple methods and datasets. We extend FCOS3D with our IP-Head (IP-FCOS3D, Figure \ref{fig:ip_networks}) as the detector in the LR3D framework.
For qualitative evaluation, we utilize FastRCNN3D with IP-Head (Figure \ref{fig:ip_networks}) as LR3D detector, to utilize 2D conditions.

\begin{figure*}[t]
    \vspace{-0.2in}
	\centering
	\includegraphics[width=1.0\linewidth]{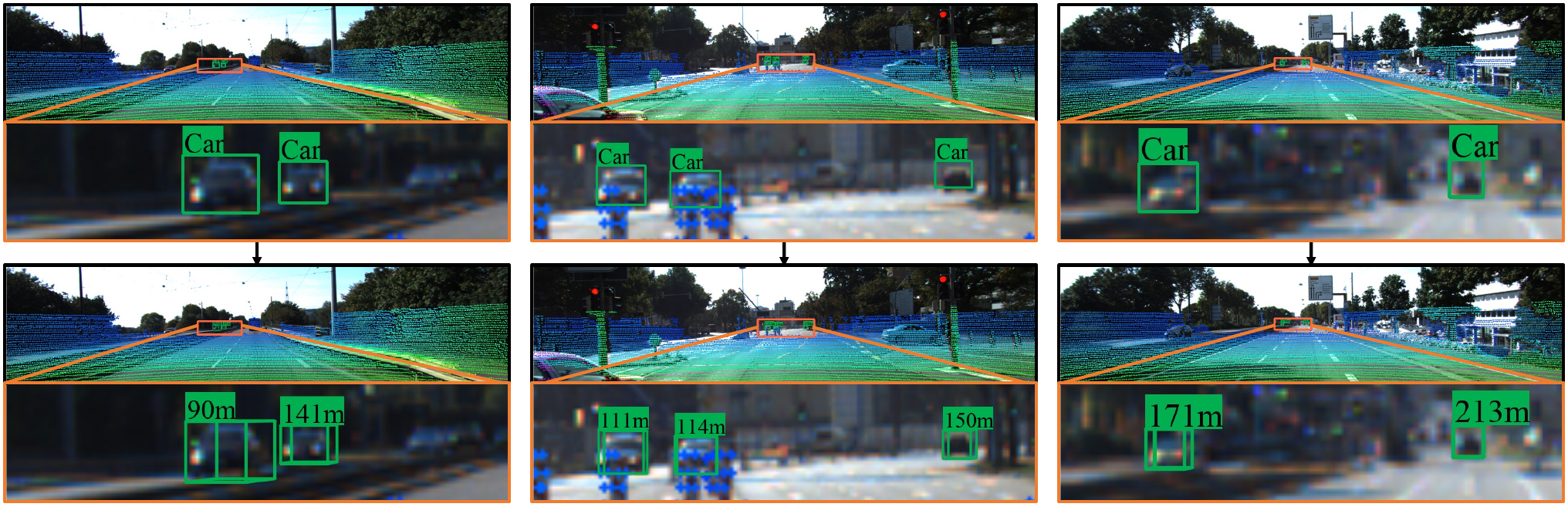}\\
	\vspace{-0.1in}
	\caption{
	Qualitative results on KITTI Dataset in detecting extremely far away 3D objects. LiDAR points are projected to images shown with different colors related to depth. The distances of 3D bounding boxes are marked on the top-left. \textbf{1st \& 2nd rows:} 2D annotations of extremely distant objects; \textbf{3rd \& 4th rows:} 3D box prediction of corresponding objects.
	}
	\label{fig:kitti_viz}
\end{figure*}
\myparagraph{Quantitative Results.}
We compare our LR3D, using IP-FCOS3D as its detector, and the baseline FCOS3D \cite{wang2021fcos3d} for distant 3D detection without 3D annotations in Table \ref{tab:kitti_main}. It is obvious that our LR3D outperforms FCOS3D on detecting distant objects by a large margin, i.e., 31.3\% and 27.7\% on LDS and mAP, respectively. Even compared to FCOS3D with distant 3D ground truth supervision, LR3D still shows competitive performance on distant objects.

We further evaluate the performance of using LR3D as the teacher model to generate pseudo long-range 3D labels to train state-of-the-art camera-based detectors. The results are listed in Table \ref{tab:kitti_main}. Surprisingly, models trained with the combination of close 3D ground truth and distant 3D pseudo labels even achieve comparable or even better performance to their fully 3D supervised counterparts. 

Quantitative results in Table \ref{tab:kitti_main} demonstrate the effectiveness of our LR3D framework, as well as the potential of our method in relieving the great demand of accurate 3D bounding box annotations for long-range object detection.

\myparagraph{Qualitative Results.}
We visualize the 3D prediction results of LR3D conditioned with ``DontCare'' objects in Figure \ref{fig:kitti_viz}. It is also clear that supervised by close 3D annotations only (maximum to 40m away), our method infers reasonable 3D bounding boxes (3rd \& 4th rows) for those extremely distant objects much beyond the 3D supervision range according to the corresponding 2D bounding boxes (1st \& 2nd rows). These qualitative results further demonstrate the potential to extend LR3D to the labeling process. Annotators can first simply label 2D boxes for distant objects and then generate associated 3D annotations by LR3D.

\begin{table*}[t]
    \vspace{-0.1in}
   \centering 
   \setlength{\tabcolsep}{2.1mm}{
   \footnotesize
   \begin{tabular}{l|c||cc||cc||cc||cc}
       \hline
       \multicolumn{1}{c|}{ \multirow{2}{*}{Method}} &
       \multicolumn{1}{c||}{ \multirow{2}{*}{\makecell[c]{Distant 3D\\Groundtruth?}}} &
       \multicolumn{2}{c||}{Overall} & 
       \multicolumn{2}{c||}{Close (0m-40m)} & 
       \multicolumn{2}{c||}{Distant (40m-51.2m)} &
       \multicolumn{2}{c}{Distant (51.2m-Inf)}\\ \cline{3-10}
       \multicolumn{1}{c|}{} & 
       \multicolumn{1}{c||}{} & 
       LDS (\%) & mAP (\%) &
       LDS (\%) & mAP (\%) &
       LDS (\%) & mAP (\%) & 
       LDS (\%) & mAP (\%) \\
       \hline
       \hline
       FCOS3D \cite{wang2021fcos3d} & $\surd$ & 26.6 & 27.4 & 29.1 & 30.1 & 16.4 & 12.7 & 11.3 & 8.6 \\
       \hline
       FCOS3D \cite{wang2021fcos3d} & \multicolumn{1}{c||}{ \multirow{2}{*}{-}} & 22.6 & 23.4 & 29.9 & 30.3 & 1.8 & 1.4 & 0.0 & 0.0 \\
       \makecell[l]{LR3D (IP-FCOS3D)} & {} & \bf 24.1 & \bf 24.4 & 28.5 & 28.4 &  \bf 16.1 & \bf 11.3 & \bf 6.4 & \bf 4.3 \\
       \hline
       \hline
       \multicolumn{10}{l}{\multirow{1}{*}{\bf Long-range Teacher}} \\
       \hline
       \hline
       BEVFormer-S \cite{li2022bevformer} & $\surd$ & 36.9 & 37.2 & 38.3 & 38.7 & 19.6 & 15.2 & - & - \\
       \hline
       BEVFormer-S \cite{li2022bevformer} & \multicolumn{1}{c||}{ \multirow{2}{*}{-}} & 33.6 & 34.2 & 38.0 & 38.5 & 2.4 & 1.7 & - & - \\
       +LR3D teacher & {} & \bf 36.3 & \bf 36.8 & 38.0 & 38.9 & \bf 18.3 & \bf 10.9 & - & - \\
       \hline
       \hline
       BEVFormer-S$^\dagger$ \cite{li2022bevformer} & $\surd$ & 33.7 & 34.0 & 37.4 & 37.9 & 19.4 & 14.5 & 14.4 & 10.2 \\
       \hline
       BEVFormer-S$^\dagger$ \cite{li2022bevformer} & \multicolumn{1}{c||}{ \multirow{2}{*}{-}} & 29.1 & 29.9 & 38.0 & 38.7 & 2.4 & 1.0 & 0.0 & 0.0 \\
       +LR3D teacher & {} & \bf 33.0 & \bf 33.6 & 37.5 & 38.0 & \bf 17.9 & \bf 10.7 & \bf 12.7 & \bf 6.4 \\
       \hline
   \end{tabular}}
   \vspace{-0.1in}
   \caption{Comparison on state-of-the-art methods with and without IP-Head or LR3D teacher supervised by distant 2D ground truth only on the nuScenes val dataset. Their fully supervised counterparts (with distant 3D ground truth) are also illustrated.
   }\label{tab:nusc_main}
   \vspace{-0.2in}
\end{table*}

\subsection{Results on nuScenes}
\myparagraph{Data Preparation.}
NuScenes \cite{nuscenes2019} is a large-scale dataset with 1,000 autonomous driving sequences labeled with 3D boxes in 10 classes. In nuScenes, we set the distance threshold of marking long-range objects as 40m.
We mark those annotations beyond the threshold as distant objects, remove the 3D annotations and only use 2D box labels for training. For those closer than 40m, we keep their both 2D/3D labels.

We evaluate different models among all 10 classes, and do not set the detection range upper-bound for different classes as the official benchmark \cite{nuscenes2019}. This setting helps keep distant predictions. Following \cite{nuscenes2019}, we train our model on 700 training scenes and test it on 150 validation scenes.

\myparagraph{Model Setting.}
We utilize FCOS3D \cite{wang2021fcos3d} as our monocular baseline and enhance it with IP-Head (IP-FCOS3D) for LR3D with 2D supervision only. To demonstrate the effectiveness of LR3D as a long-range teacher on nuScenes, we choose BEVFormer \cite{li2022bevformer} as the student for its effectiveness, and test its performance without distant 3D supervision.

\myparagraph{Main Results.}
The experimental results are illustrated in Table \ref{tab:nusc_main}. With IP-Head, LR3D outperforms its baseline  FCOS3D by 14.3\% and 9.9\% performance improvement in terms of LDS and mAP for distant objects from 40m to 51.2m. It is significant. Compared to the fully 3D supervised FCOS3D with distant 3D ground truth, LR3D still yields competitive capacity on long-range 3D detection.

We also list the results of BEVFormer \cite{li2022bevformer} with and without LR3D teacher on detecting distant 3D objects without 3D supervision. As illustrated, with the LR3D teacher, BEVFormer-S yields strong performance, comparable to its fully-supervised counterparts with 3D box labels. 

Given the limited perception range of official BEVFormer (-51.2m to 51.2m), we enlarge the perception range of BEVFormer from 51.2m to 76.8m for further verification, denoted as BEVFormer-S$^\dagger$. Still, as shown in Table \ref{tab:nusc_main}, with LR3D teacher, BEVFormer-S$^\dagger$ achieves impressive improvement compared to its baseline, and yields comparable results to those of the fully-supervised model.

\begin{table*}[t]
\captionsetup[subfloat]{aboveskip=2pt,belowskip=1pt}
\vspace{-.2in}
\centering
\subfloat[
Effect of parameter learning in $f^{(\theta)}$.
\label{tab:ablation_iphead}
]
{
\centering
\begin{minipage}{0.29\linewidth}{\begin{center}
\tablestyle{1.2pt}{1.05}
\begin{tabular}{c|c|c|c}
		\hline
		\multicolumn{1}{c|}{ \multirow{3}{*}{\makecell[c]{$f^{(\theta)}$ Weight \\ Learning \\ Strategy}}} & 
		\multicolumn{1}{c|}{ \multirow{2}{*}{\makecell[c]{Overall \\ (0m-Inf)}}} &
		\multicolumn{1}{c|}{ 
		\multirow{2}{*}{\makecell[c]{Close \\ (0m-40m)}}} &
		\multicolumn{1}{c}{ \multirow{2}{*}{\makecell[c]{Distant \\ (40m-Inf)}}} \\
		& & &  \\
		& LDS (\%) & LDS (\%)& LDS (\%) \\
		\hline
		Shared & 45.0 & 46.6 & 32.5 \\
		\rowcolor{LightCyan} Dynamic & \bf 50.0 & \bf 52.1 & \bf 36.2 \\
		\hline
		\multicolumn{4}{c}{~} \\
	\end{tabular}
\end{center}}\end{minipage}
}
\hspace{2em}
\subfloat[
Effect of positional encoding in $f^{(\theta)}$.
\label{tab:ablation_posenc}
]{
\centering
\begin{minipage}{0.29\linewidth}{\begin{center}
\tablestyle{1.2pt}{1.05}
\begin{tabular}{ l | c | c | c }
		\hline
		\multicolumn{1}{c|}{ \multirow{3}{*}{\makecell[c]{Pos. Enc.}}} & 
		\multicolumn{1}{c|}{ \multirow{2}{*}{\makecell[c]{Overall \\ (0m-Inf)}}} &
		\multicolumn{1}{c|}{ \multirow{2}{*}{\makecell[c]{Close \\ (0m-40m)}}} &
		\multicolumn{1}{c}{ \multirow{2}{*}{\makecell[c]{Distant \\ (40m-Inf)}}} \\
		& & &  \\
		& LDS (\%) & LDS (\%)& LDS (\%) \\
		\hline
		None & 19.9 & 20.2 & 13.3  \\
		\rowcolor{LightCyan} Sin. Cos. \cite{AttentionIsAllYouNeed} & \bf 50.0 & \bf 52.1 & \bf 36.2 \\
		\hline
		\multicolumn{4}{c}{~} \\
	\end{tabular}
\end{center}}\end{minipage}
}
\hspace{2em}
\subfloat[
Effect of 2D box descriptors.
\label{tab:ablation_ipf2d}
]{
\centering
\begin{minipage}{0.29\linewidth}{\begin{center}
\tablestyle{1.2pt}{1.05}
\begin{tabular}{ l | c | c | c }
		\hline
		\multicolumn{1}{c|}{ \multirow{3}{*}{\makecell[c]{2D Box ($b_{2d}$) \\ Descriptors}}} & 
		\multicolumn{1}{c|}{ \multirow{2}{*}{\makecell[c]{Overall \\ (0m-Inf)}}} &
		\multicolumn{1}{c|}{ \multirow{2}{*}{\makecell[c]{Close \\ (0m-40m)}}} &
		\multicolumn{1}{c}{ \multirow{2}{*}{\makecell[c]{Distant \\ (40m-Inf)}}} \\
		& & &  \\
		& LDS (\%) & LDS (\%)& LDS (\%) \\
		\hline
		box width ($w_{2d}$) & 48.5 & 51.7 & 30.6 \\
		box height ($h_{2d}$) & 49.4 & 51.8 & 34.9 \\
		\rowcolor{LightCyan} both ($w_{2d}$, $h_{2d}$) & \bf 50.0 & \bf 52.1 & \bf 36.2 \\
		\hline
	\end{tabular}
\end{center}}\end{minipage}
}
\\
\centering
\vspace{.2em}
\subfloat[
Effect of the number of layers in $f^{(\theta)}$.
\label{tab:ablation_ipf_layer}
]{
\begin{minipage}{0.29\linewidth}{\begin{center}
\tablestyle{4pt}{1.05}
\begin{tabular}{ c | c | c | c }
		\hline
		\multicolumn{1}{c|}{ \multirow{3}{*}{\makecell[c]{No. of \\ layers}}} & 
		\multicolumn{1}{c|}{ \multirow{2}{*}{\makecell[c]{Overall \\ (0m-Inf)}}} &
		\multicolumn{1}{c|}{ \multirow{2}{*}{\makecell[c]{Close \\ (0m-40m)}}} &
		\multicolumn{1}{c}{ \multirow{2}{*}{\makecell[c]{Distant \\ (40m-Inf)}}} \\
		& & &  \\
		& LDS (\%) & LDS (\%)& LDS (\%) \\
		\hline
		1 & 48.6 & 51.3 & 32.4 \\
		\rowcolor{LightCyan} 2 & \bf 50.0 & \bf 52.1 & \bf 36.2 \\
		3 & 49.7 & 52.0 & 35.2 \\
		4 & 49.0 & 51.4 & 34.9 \\
		\hline
	\end{tabular}
\end{center}}\end{minipage}
}
\hspace{2em}
\subfloat[
Effect of the channel number in $f^{(\theta)}$.
\label{tab:ablation_ipf_channel}
]{
\begin{minipage}{0.29\linewidth}{\begin{center}
\tablestyle{4pt}{1.05}
\begin{tabular}{ c | c | c | c }
		\hline
		\multicolumn{1}{c|}{ \multirow{3}{*}{\makecell[c]{No. of \\ channels}}} & 
		\multicolumn{1}{c|}{ \multirow{2}{*}{\makecell[c]{Overall \\ (0m-Inf)}}} &
		\multicolumn{1}{c|}{ \multirow{2}{*}{\makecell[c]{Close \\ (0m-40m)}}} &
		\multicolumn{1}{c}{ \multirow{2}{*}{\makecell[c]{Distant \\ (40m-Inf)}}} \\
		& & &  \\
		& LDS (\%) & LDS (\%)& LDS (\%) \\
		\hline
		8 & 48.4 & 50.9 & 34.5 \\
		\rowcolor{LightCyan} 16 & \bf 50.0 & 52.1 & \bf 36.2 \\
		32 & 49.9 & \bf 52.3 & 35.6 \\
		64 & 49.7 & 51.9 & 36.1 \\
		\hline
	\end{tabular}
\end{center}}\end{minipage}
}
\hspace{2em}
\subfloat[
Effect of the projection augmentation.
\label{tab:ablation_projaug}
]{
\begin{minipage}{0.29\linewidth}{\begin{center}
\tablestyle{2pt}{1.05}
\begin{tabular}{ l | c | c | c }
		\hline
		\multicolumn{1}{c|}{ \multirow{3}{*}{\makecell[c]{Aug. \\ Strategy}}} & 
		\multicolumn{1}{c|}{ \multirow{2}{*}{\makecell[c]{Overall \\ (0m-Inf)}}} &
		\multicolumn{1}{c|}{ \multirow{2}{*}{\makecell[c]{Close \\ (0m-40m)}}} &
		\multicolumn{1}{c}{ \multirow{2}{*}{\makecell[c]{Distant \\ (40m-Inf)}}} \\
		& & &  \\
		& LDS (\%) & LDS (\%)& LDS (\%) \\
		\hline
		FCOS3D \cite{wang2021fcos3d} & 42.1 & 50.5 & 4.9 \\
		+Copy-Paste & N/A & N/A & N/A \\
		\hline
		LR3D & 48.3 & \bf 52.7 & 26.5 \\
		\rowcolor{LightCyan} +Proj. Aug. & \bf 50.0 & 52.1 & \bf 36.2 \\
		\hline
	\end{tabular}
\end{center}}\end{minipage}
}

\vspace{-0.1in}
\caption{Ablation studies on IP-Head structure and projection augmentation. Default settings are highlighted in \colorbox{LightCyan}{lightcyan}.}
\label{tab:ablations} 
\vspace{-0.2in}
\end{table*}
\subsection{Ablation Studies}
All ablation studies are conducted on KITTI Dataset, using IP-FCOS3D as the detector of the LR3D model.

\myparagraph{Analysis on the IP-Head.}
IP-Head learns to estimate implicit inverse function $f^{-1}$, which is the mapping from 2D bounding box on image to corresponding depth, through two MLP networks $f_g$ and $f^{(\theta)}$. We evaluate its effectiveness by comparing the mapping curve from 2D box size to the depth of a fixed 3D object generated by ground truth transformation and estimated transformation. 

\begin{figure}[t]
  \centering
  \includegraphics[width=1.0\linewidth]{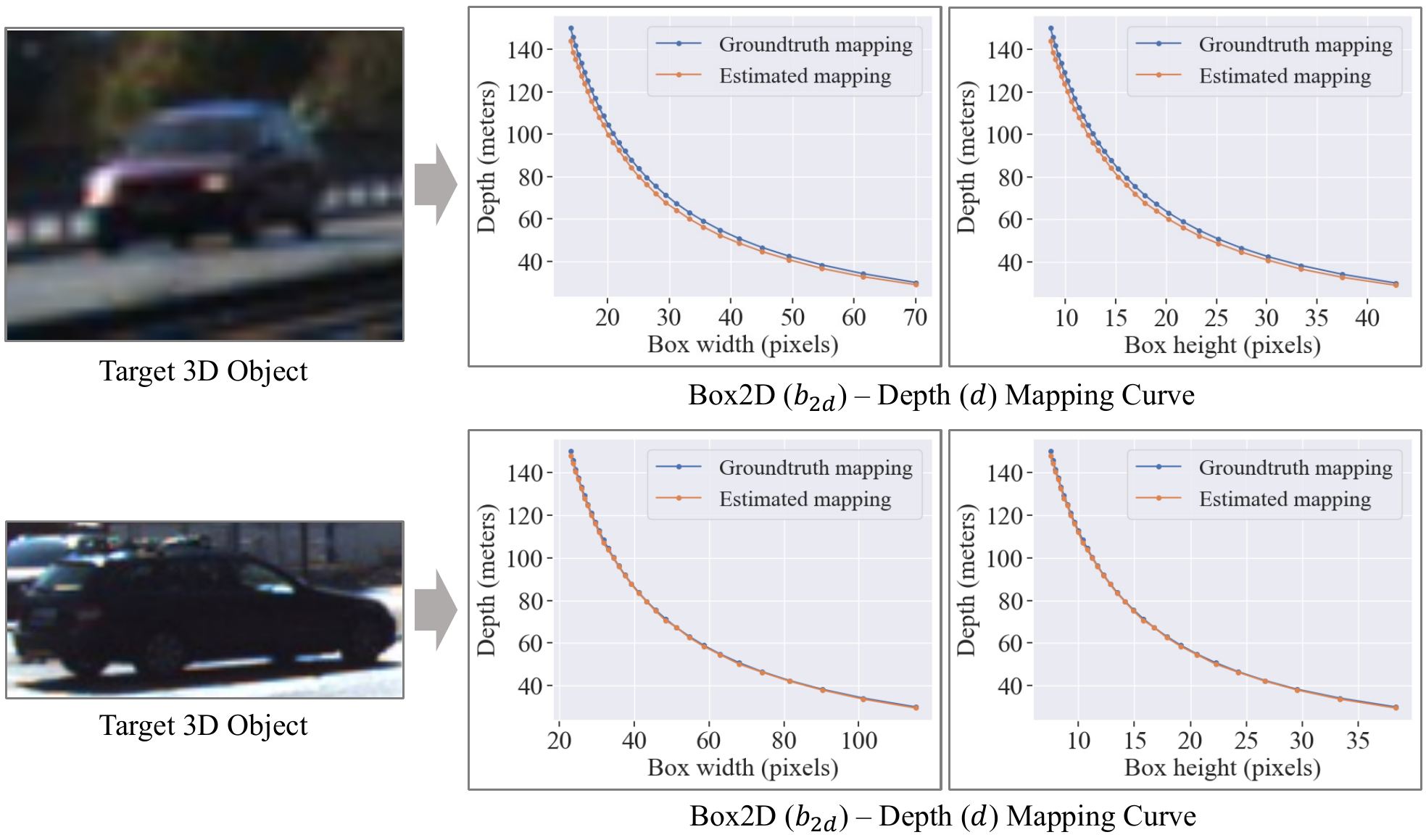}\\
  \vspace{-0.1in}
  \caption{Illustration of the ground truth and estimated $b_{2d}$-$d$ mappings. Each row indicates the target 3D object and its mapping from 2D box width and height to the depth.}
  \label{fig:ip_curve}
  \vspace{-0.2in}
\end{figure}

The generation of the ground truth curve follows Eq. \eqref{eq:ipl_f}, from which we obtain multiple $d$-$b_{2d}$ pairs by modifying depth $d$. The estimated curve is generated through Eq. \eqref{eq:ipl_f_inverse_sim} by changing the 2D box inputs. As illustrated in Figure \ref{fig:ip_curve}, our IP-Head can estimate specific implicit inverse functions for different objects (by comparison across rows). The estimated function by MLP network $f^{(\theta)}$ well models the mapping from 2D bounding box size to the corresponding depth of target 3D object (through comparison inside each row). 

We further provide quantitative analysis of using MLP network $f_g$ to dynamically determine weights of $f^{(\theta)}$ to estimate specific implicit inverse functions of different 3D objects. As listed in Table \ref{tab:ablation_iphead}, compared to the design of shared weights, using dynamic weights assignment in $f^{(\theta)}$ well models the difference of implicit inverse functions due to size and orientation of 3D objects, and thus outperforms the model using a shared weight $\theta$ across 3D objects.

\myparagraph{Analysis on the IP-Head Structure.}
We first describe the default structure of IP-Head. $f^{(\theta)}$ is a 2-layer perceptron with channels of 16 and 1. The 2D boxes $b_{2d}$ are described as height ($h_{2d}$) and width ($w_{2d}$). The positional encoding function $f_{\text{PE}}$ is the sine-cosine function illustrated in \cite{AttentionIsAllYouNeed}.

In Table \ref{tab:ablation_posenc}, we evaluate the contribution of positional encoding. With positional encoding \cite{AttentionIsAllYouNeed}, the 2D descriptors are well encoded into more informative features, which thus outperforms the model without positional encoding (``None''). In Table \ref{tab:ablation_ipf2d}, we compare the model with different 2D box descriptors. Utilizing both 2D box width $w_{2d}$ and height $h_{2d}$ as descriptors yields the best performance. In Table \ref{tab:ablation_ipf_layer} and Table \ref{tab:ablation_ipf_channel}, we conduct ablation studies on MLP structures of network $f^{(\theta)}$ with different numbers of layers and channels. As illustrated, the model with 2-layer and channel number 16 achieves the best results.

\myparagraph{Analysis on the Projection Augmentation.}
The projection augmentation ensures the estimated implicit inverse function $f^{(\theta)}$ models the relation between 2D box sizes and depths by generating more training $b_{2d}$-$d$ pairs. We analyze it by comparing the performance of LR3D trained with and without this augmentation. As shown in Table \ref{tab:ablation_projaug}, compared to FCOS3D baseline (1st row), LR3D can model mapping functions from 2D sizes to depths and produces good results on distant objects without 3D supervision. Moreover, equipped with projection augmentation, LR3D further shows a huge performance improvement on distant objects.

A similar augmentation to projection augmentation is copy-paste augmentation \cite{mmcopypaste2020zhang,wang2021pointaugmenting}, which simulates distant training samples by cropping close objects with image patches and pasting them to distant areas by Eq. \eqref{eq:ipl_f}. However, this augmentation brings noises in image distribution, is hard to optimize, and thus leads to a great performance drop (Table \ref{tab:ablation_projaug} (2nd row)). In contrast, projection augmentation is simple to implement without changing training images, and able to augment to any depth without extra effort.

\subsection{Results on Other Datasets}

Please find additional results on Cityscapes3D \cite{cityscape3d}, Waymo \cite{Waymo}, and Argoverse 2 \cite{Argoverse2} in the supplementary materials.
\section{Conclusion}
In this paper, we proposed the LR3D framework for long-range 3D detection using 2D supervision and designed the IP-Head as the key component. We developed a teacher-student pipeline to benefit all camera-based detectors. Our LDS, a metric with relative distance criterion, helps obtain informative quantitative results for distant objects. With these designs, we demonstrate the feasibility of using 2D supervision only for long-range 3D detection.

{
    \small
    \bibliographystyle{unsrt}
    \bibliography{main}
}


\end{document}